\documentclass[final]{cvpr}

\usepackage{nopageno}
\usepackage{times}
\usepackage{epsfig}
\usepackage{graphicx}
\usepackage{amsmath}
\usepackage{amssymb}
\usepackage{color}
\usepackage{multirow}
\usepackage{comment}
\usepackage{enumitem}
\usepackage{float}

\usepackage[pagebackref=true,breaklinks=true,colorlinks,bookmarks=false]{hyperref}

\def\cvprPaperID{9917} %
\def\confYear{CVPR 2021}

\newcommand{\todo}[1]{}
\renewcommand{\todo}[1]{{\color{blue} \bf {#1}}}

\begin{document}

\title{Representation Learning via Global Temporal Alignment and Cycle-Consistency}

\author{Isma Hadji, Konstantinos G.\ Derpanis, Allan D.\ Jepson\\
Samsung AI Centre Toronto\\
{\tt\small \{isma.hadji, allan.jepson\}@samsung.com \hspace{10pt} k.derpanis@partner.samsung.com}\\
}

\maketitle

\begin{abstract}
We introduce a weakly supervised method for representation learning based on aligning temporal sequences (e.g., videos) of the same process (e.g., human action). 
The main idea is to use the global temporal ordering of latent correspondences %
across sequence pairs as a supervisory signal.  In particular,
we propose a loss based on scoring the 
optimal sequence alignment to train an embedding network.  %
Our loss is based on a novel probabilistic path finding view of dynamic time warping (DTW) that contains the following three key features: (i)
the local path routing decisions are contrastive  and differentiable,
(ii) pairwise distances are cast as probabilities that are contrastive as well, and (iii) 
our formulation naturally admits 
a global cycle-consistency loss that verifies correspondences.
For evaluation, we consider the tasks of fine-grained action classification, few shot learning, and video synchronization. %
We report significant performance increases over previous methods. In addition, we report two applications of our temporal alignment framework, namely 3D pose reconstruction and fine-grained audio/visual retrieval.
\end{abstract}

\section{Introduction}
\begin{figure}[t]
\centering
  \includegraphics[width=\linewidth]{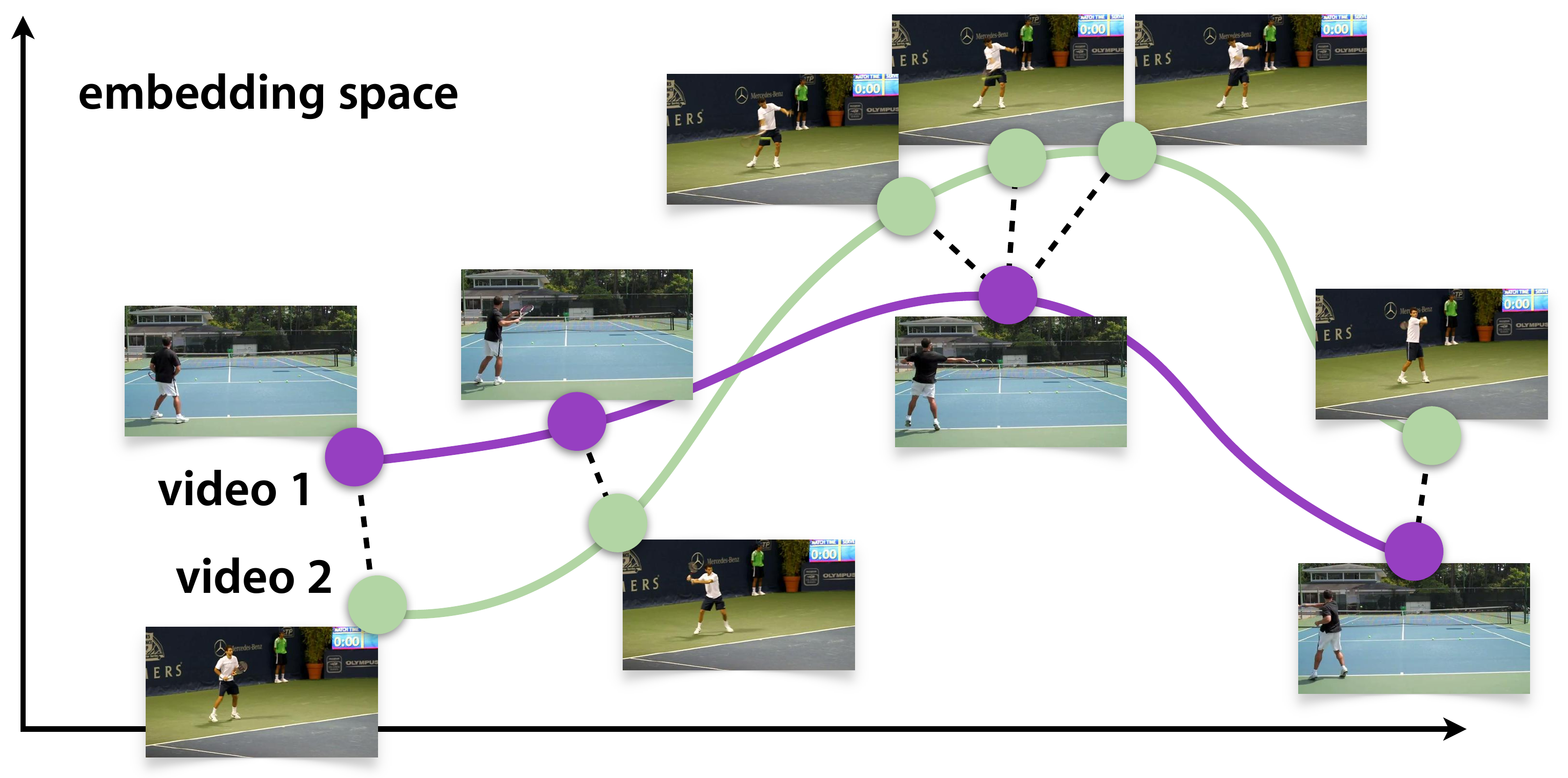}%
\caption{We introduce a representation learning approach based on (globally) aligning pairs of temporal sequences (\eg, video) depicting the same process (\eg, human action).
Our training objective is to learn an element-wise embedding function that supports the alignment process.
  For example, here we illustrate the alignment (denoted by black dashed lines) in the embedding space
between videos of the same human action (\ie, tennis forehand) containing significant variations in their appearances and dynamics. %
  Empirically, we show that our learned embeddings are sensitive to both human pose and fine-grained temporal distinctions, while being invariant to appearance, camera viewpoint, and background.}
\label{teaser}
\vspace{-5pt}
\end{figure}

Temporal sequences (\eg, videos) are an appealing data source as they provide a rich %
source of information and additional constraints to leverage in learning.
By far the main focus on temporal sequence analysis has been on learning representations targeting distinctions at the global signal level, \eg, action classification, where abundant labeled 
data is available for training.  In this paper, we target a weakly-supervised training regime for representation learning, capable of making fine-grained temporal distinctions.

Most previous approaches to temporally fine-grained understanding of sequential signals have considered fully-supervised training methods (\eg, \cite{YeungRJAMF18}), where labels
are provided at the sub-sequence level, \eg, frames.  The major drawback of these methods is the expense in acquiring dense labels, and their subjective nature.
In contrast, a key consideration in our work is the selection of training signals
capable of scaling up to large amounts of data yet supporting finer-grained video understanding.

As outlined in Fig.\ \ref{teaser}, given a set of paired sequences capturing the same process (\eg, tennis forehand) but highly varied (\ie,
different participants, action executions, scenes, and camera viewpoints),
our method trains an embedding network to support the recovery of their latent temporal alignment.
We refer to our method as weakly supervised, as only readily available sequence-level labels (\eg, tennis forehand) are required to construct training pairings containing the same process. Given such a pair of sequences, we use their latent temporal alignment as a supervisory signal to learn fine-grained temporal distinctions.

Key to our proposed method is a novel dynamic time warping (DTW) formulation to score global alignments between paired sequences.
DTW enforces a stronger constraint over simply considering local (soft) nearest neighbour correspondences \cite{DwibediATSZ19}, since the temporal ordering of the matches in the sequences are taken into account. 
We depart from previous differentiable DTW methods \cite{MenschB18,ChangHS0N19,CaoJCCN20} by taking a probabilistic path finding view of DTW that encompasses the following three key features.
First, we introduce a 
differentiable smoothMin operator that effectively selects each successive path extension. Moreover,  we show that this operator has a contrastive effect across paths which is missing
in previous differentiable DTW formulations  \cite{MenschB18,ChangHS0N19,CaoJCCN20}. 
Second, the pairwise embedding similarities that form our cost function are defined as probabilities, using the softmax operator. 
Optimizing our loss is shown to correspond to finding the maximum probability of any feasible alignment between the paired sequences.
The softmax operator over element pairs also provides a contrastive component which we show is crucial to prevent the model from learning trivial embeddings.
This forgoes the need for a downstream discriminative loss and the corresponding 
non-trivial task of defining negative alignments, \eg, \cite{ChangHS0N19,CaoJCCN20}. 
Third, as an additional supervisory signal, our probabilistic framework admits a straightforward
global cycle-consistency loss that matches the alignments recovered through a cycle of sequence pairings.
Collectively, our method takes into account long-term temporal information that allows 
us to learn embeddings sensitive to 
fine-grained temporal distinctions (\eg, human pose),
while being invariant to nuisance variables, \eg, camera viewpoint, background, and appearance.%

\vspace{5pt}
\noindent{\bf Contributions. } We make the following key contributions:   
\begin{itemize}[topsep=0ex,itemsep=-1ex,partopsep=0ex,parsep=1ex]
    \item A novel weakly supervised method for representation
learning tasked with discovering the alignment between sequence pairings for the purpose of fine-grained temporal understanding.
    \item A differentiable DTW formulation with two novel features: (i) a smoothMin operation that admits a probabilistic path interpretation and is contrastive across alternative paths, and (ii) a probabilistic data term that is contrastive across alternative data pairs.
    \item A global cycle consistency loss to further enforce the temporal alignment.
    \item %
    An extensive set of evaluations, ablations, and comparisons with previous methods. We report significant performance increases on several tasks requiring fine-grained temporal distinctions.
    \item Two downstream applications, namely 3D pose reconstruction and audio-visual retrieval.
\end{itemize}
Our code and trained models will be available at: \href{https://github.com/hadjisma/VideoAlignment}{https://github.com/hadjisma/VideoAlignment}.

\section{Related work}

\noindent{\bf Representation learning. }
Most focus in representation learning with videos has been cast in a fully supervised setting, \eg, \cite{TranBFTP15,SimonyanZ14,CarreiraZ17,Feichtenhofer0M19}.
Self-supervised learning with images or videos has emerged 
as a viable alternative to supervised learning, where the supervisory signal is obtained from the data.
For video, a variety of proxy tasks have been defined in lieu of training with annotations, such as classifying whether video frames are in the correct temporal 
order (\eg, \cite{MisraZH16,FernandoBGG17,LeeHS017,BuchlerBO18,Xu0ZSXZ19,WeiLZF18}), predicting whether a video is played at a normal or modified rate \cite{BenaimELMFRID20}, solving a spatiotemporal jigsaw puzzle task \cite{AhsanME19}, predicting figure-ground segmentation \cite{PathakGDDH17}, predicting pixel \cite{YuHD16,MeisterH018,VondrickSFGM18,JanaiGRBG18} or region correspondences \cite{WangG15,WangJE19,abs-2006-14613} across neighbouring video frames,
and predicting some aspect of future frames conditioned on past frames \cite{VondrickPT16,WangJBHLL19,HanXZ19}.
Others have considered multimodal settings, such as predicting
video-audio misalignment \cite{OwensE18}.
Similar to \cite{DwibediATSZ19}, our method is best characterized as weakly supervised, where sequence-level labels are used to determine sequence pairings for training.

\noindent{\bf Sequence alignment. } Several methods \cite{SigurdssonGSFA18,SermanetLCHJSLB18} assume paired, temporally synchronized videos of the same physical event for the purpose of representation learning.  
In contrast, and more closely related to our work, are methods that seek the alignment between sequences capturing the same process.  One approach \cite{DwibediATSZ19} is to cast the learning objective as maximizing the number of elements between sequences that can be brought into one-to-one correspondence via (soft) nearest neighbours.  This method does not leverage the long-term temporal structure of the sequences as done in 
dynamic time warping (DTW) \cite{saoke78}.  Given a cost function, DTW finds the optimal alignment between two sequences defined between elements comprising the sequences.  Recent efforts \cite{CuturiB17,MenschB18,CaiXYHR19,ChangHS0N19,CaoJCCN20} have explored differentiable approximations of the discrete operations underlying DTW to allow gradient-based training.  Similar to recent work \cite{ChangHS0N19,CaoJCCN20}, we also incorporate a relaxed DTW as our loss for sequence alignment.  
Our formulation is probabilistic and includes a 
contrastive definition of the element-wise similarities.
A key distinction with these prior works and our own, beyond differences in the target application domain (\eg, video-transcript alignment \cite{ChangHS0N19}), is that rather than incorporate contrastive modelling after the DTW step (\eg, through the use of margin-based loss),
our method includes contrastive signals in both the differentiable min approximation and the pairwise matching cost function used in our DTW framework. %

\noindent{\bf Contrastive learning.} %
One can also draw parallels with contrastive learning 
using the cross-entropy loss (\ie, negative log softmax) \cite{GutmannH10,MnihK13,abs-1807-03748},
where the goal is to learn a representation that brings
different views of the same data together (\ie, positives) in the embedding space, while pushing views of different data (\ie, negatives) apart.
This amounts to encoding information shared across the views, while eschewing unique factors to each view.
To construct different views, previous work has explored a variety of augmentation and sampling schemes  \cite{DosovitskiyFSRB16,abs-1906-05849,chen2020,HanXZ19,abs-2003-07990,He0WXG20} 
and correspondences across different modalities (\eg, video-audio, video-text, and luminance-depth \cite{ChungCK19,abs-1906-05849,abs-1906-05743,abs-2003-04298,MiechASLSZ20}).  In these works, the positive and negative pairings are known by construction, \eg, via image augmentation.
We also make use of contrastive losses, but note that the correspondences (\ie, positives) between the sequences are latent rather  than known.

\noindent{\bf Cycle-consistency. } In addition to alignment, our method also incorporates cycle-consistency as a supervisory signal, where the objective is to verify matches 
across sets.  
Similar to recent work \cite{DwibediATSZ19}, we apply cycle-consistency across two temporal sequences.  A 
key difference is that previous work \cite{DwibediATSZ19} applies cycle-consistency independently to local matches across sequences; whereas, we consider both local matches and their global temporal ordering, which we demonstrate empirically leads to improved alignments.

\begin{figure*}[t]
\centering
  \includegraphics[width=0.9\linewidth]{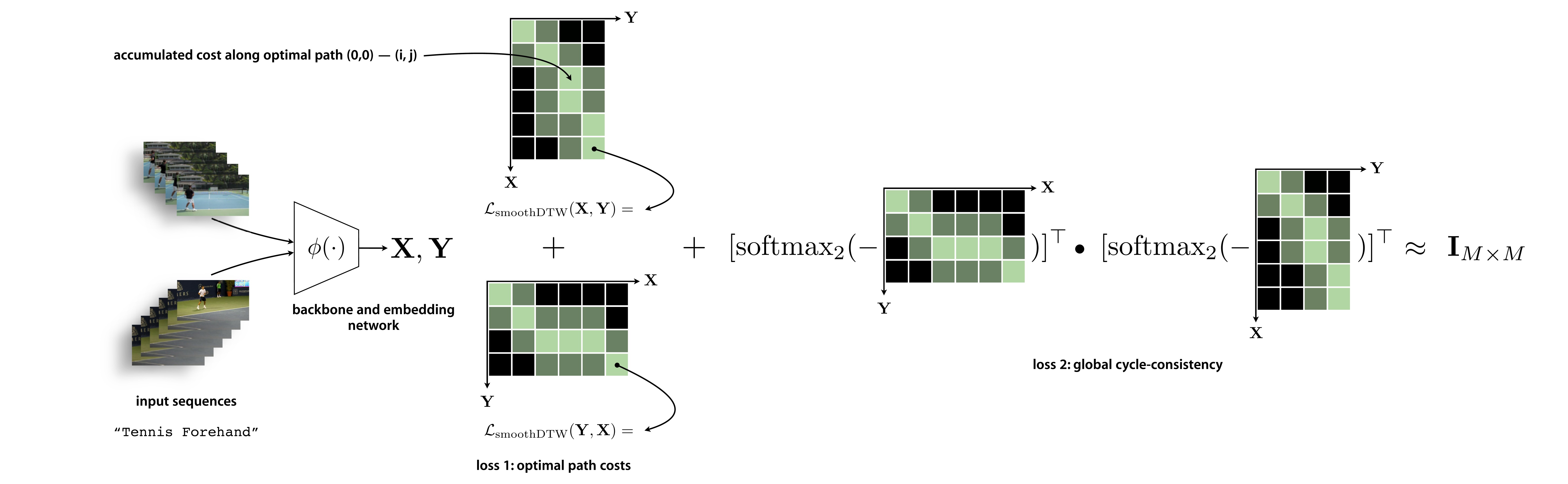}%
\caption{Our sequence alignment approach to representation learning begins by encoding each element comprising our sequences (\eg, image frames) using a 
 trainable framewise backbone encoder plus embedding network, $\phi(\cdot)$, yielding two sequences of embeddings,  
 $\mathbf{X}$ and  $\mathbf{Y}$.
The cost of matching these two sequences is expressed as negative log probabilities and consists of two parts: (i) alignment losses, $\text{smoothDTW}(\cdot, \cdot)$, from $\mathbf{X} \mbox{ to } \mathbf{Y}$ and $\mathbf{Y} \mbox{ to } \mathbf{X}$ based on 
the cumulative cost along the optimal respective paths
and (ii)
a global cyclic-consistency loss that verifies the correspondences computed 
between each ordered pair of sequences,
where $\cdot$ denotes matrix multiplication
and $\mathbf{I}_{M\times M}$ is the square identity matrix. 
Note, our alignment cost $\text{smoothDTW}(\cdot, \cdot)$ is not symmetric in its two arguments (due to the pairwise matching cost in (\ref{eq:contrastive_loss})).  
Higher intensities in the cells comprising the accumulated cost matrices indicate lower values.}
\label{fig:loss}
\vspace{-15pt}
\end{figure*}

\section{Technical approach}

In this section, we describe our weakly-supervised approach to representation learning based on the alignment of sets of sequence pairs that capture the same process.  %
Our learning objective is the training of a shared embedding function applied to each sequence element. %
In the case of multimodal sequences (\eg, audio-video), we have separate encoders for each modality that map
their inputs to a common embedding space. 
Figure \ref{fig:loss} provides an illustrative overview of our alignment approach to representation learning, which we fully unpack in the following subsections.

\subsection{Background} \label{sec:background}
Dynamic time warping (DTW) computes the optimal alignment between two sequences, $\mathbf{X}$ and $\mathbf{Y}$, subject to certain
alignment constraints.  Let $\mathbf{X} = \begin{bmatrix} \mathbf{x}_1 &  \mathbf{x}_2 & \dotsb  &  \mathbf{x}_M \end{bmatrix} \in \mathbb{R}^{D\times M}$
and  $\mathbf{Y} \in \mathbb{R}^{D\times N}$ denote the two sequences, where $D$ corresponds to the dimensionality
of the constituent sequence elements and $M$, $N$ the respective sequence lengths.  Given a cost matrix, $\mathbf{C}  \in \mathbb{R}^{M\times N}$,
with elements $c_{i,j}$ defined by a cost function, $c(\mathbf{x}_i, \mathbf{y}_j)$,
that measures the cost of matching elements  $\mathbf{x}_i$ and $\mathbf{y}_j$, we seek a feasible path between $c_{0,0}$ and
$c_{M,N}$ that minimizes the total accumulated cost. 
The feasible paths are subject to
matching endpoints, monotonicity, and continuity constraints. %
While not used here, the endpoint constraint can be relaxed to allow for subsequence matching, \eg, \cite{SakuraiFY07,CaoJCCN20}.

The number of feasible alignments in DTW grows exponentially with the sequence lengths. Fortunately, the structure of DTW with an appropriate cost function,  $c(\mathbf{x}_i, \mathbf{y}_j)$, is amenable to dynamic programming \cite{bellman1952} which shares both quadratic time and space complexity.  The optimal 
alignment (\ie, path through the cost matrix) is found by evaluating the following recurrence \cite{saoke78}:
\begin{align}
   \mathbf{R}&(i, j) =  c(\mathbf{x}_i, \mathbf{y}_j)\hspace{3pt} +   \label{DTW:recurr} \\
   &\min({\small \begin{bmatrix}\mathbf{R}(i-1, j-1) &  \mathbf{R}(i-1, j) & \mathbf{R}(i, j-1)\end{bmatrix}^\top}), \nonumber
\end{align}
where $\mathbf{R}(0, 0) = 0$, $\mathbf{R}(0, :) = \mathbf{R}(:, 0) = \infty$, and $\mathbf{R}(i, j)$ stores the partial accumulated cost 
along the optimal and feasible path ending with the alignment between $\mathbf{x}_i$ and  $\mathbf{y}_j$.
The minimum operation amounts to a first-order Markov assumption, where the local path routing is deterministic.

Due to the discrete nature of the $\text{min}$ operator in (\ref{DTW:recurr}), responsible for local correspondence decisions, several works \cite{CuturiB17,ChangHS0N19,CaoJCCN20}
 have considered smooth variants
suitable for gradient-based training.  In Sec.\ \ref{sec:smooth-min}, we introduce a smooth relaxation of the $\text{min}$ operator 
with favourable properties for our representation learning setting.  Then in Sec.\ \ref{sec:contrastive_cost}, we define our cost
function which introduces a contrastive learning signal throughout the alignment process.

\subsection{Local differentiable decisions} \label{sec:smooth-min}
For brevity we use the notation 
\begin{equation}
\mathbf{r}_{i,j} = {\small \begin{bmatrix} \mathbf{R}(i-1, j-1)&  \mathbf{R}(i-1, j) & \mathbf{R}(i, j-1)\end{bmatrix}^\top}
\end{equation} 
to denote the incoming optimal accumulated costs from the feasible paths leading into $(i, j)$.  We first modify (\ref{DTW:recurr}) as
\begin{equation}
   \mathbf{R}(i, j) =   c(\mathbf{x}_i, \mathbf{y}_j) + \underbrace{\left[s(\mathbf{r}_{i,j}) - \min(\mathbf{r}_{i,j}) \right]}_{d(\mathbf{r}_{i,j})} +  %
    \min(\mathbf{r}_{i,j}), \label{eq:softDTWstep0}
\end{equation}
where $s(\mathbf{r}_{i,j})$ is a smooth approximation of the minimum operator.  The term $d(\mathbf{r}_{i,j})$ 
can be seen as a (non-differentiable) additional penalty on any path that reaches $(i,j)$. 
With this added penalty term, (\ref{eq:softDTWstep0}) reduces to simply
\begin{equation}
   \mathbf{R}(i, j) = c(\mathbf{x}_i, \mathbf{y}_j)\hspace{3pt} + s(\mathbf{r}_{i,j}). \label{eq:softDTW}
\end{equation}
For an appropriate choice of $s(\cdot)$ %
the right hand side is now differentiable.  Note that any path that is optimal according to (\ref{eq:softDTW}) will correspond to a feasible path for the original DTW problem (although perhaps not optimal for that problem).  Moreover, the cost of such a path according to (\ref{eq:softDTW}) will be the original cost plus the sum of the penalties $d(\mathbf{r}_{i,j})$ over all points $(i, j)$ on that path.  

We are left with choosing a smooth approximation $s(\cdot)$  %
for the minimum operator.
Here, we use a standard relaxation of the $\text{min}$ operator \cite{LangeZHV14} (specifically, the expected value $E_{i\sim q(i)}[ a_i ]$ for $q(i) := \text{softmax}( \{-a_i/\gamma\})$):
\begin{align}
\text{smoothMin}(\mathbf{a}; \gamma) =  
\left\{
\begin{matrix}
   \min\{a_i ~|~ 1 \leq i \leq N \}, & \gamma = 0\\
   \frac{\sum_{i=1}^N a_i e^{-a_i/\gamma}} {\sum_{j=1}^N e^{-a_j/\gamma}},  &  \gamma > 0
\end{matrix}\right., \label{eq:smoothmin}
\end{align}
where $\gamma$ denotes a temperature hyper-parameter. We refer to solving the recurrence relation (\ref{eq:softDTW}), with the function $s(\cdot)$ taken to be $\text{smoothMin}$, as the smoothDTW problem.

Previous alignment methods \cite{CuturiB17,ChangHS0N19,CaoJCCN20} have instead used the following $\text{min}^\gamma$ formulation as a continuous approximation of the min operator:
\begin{align}
\text{min}^\gamma(\mathbf{a}; \gamma) = 
\left\{
\begin{matrix}
   \min\{a_i~|~ 1 \leq i \leq N \}, & \gamma = 0\\
   -\gamma \log \sum_{i=1}^N e^{-a_i/\gamma} ,  &  \gamma > 0
\end{matrix}\right., \label{eq:min_gamma}
\end{align}
where again $\gamma$ denotes a temperature hyper-parameter.
\begin{figure}[t]
\centering
  \includegraphics[width=2in]{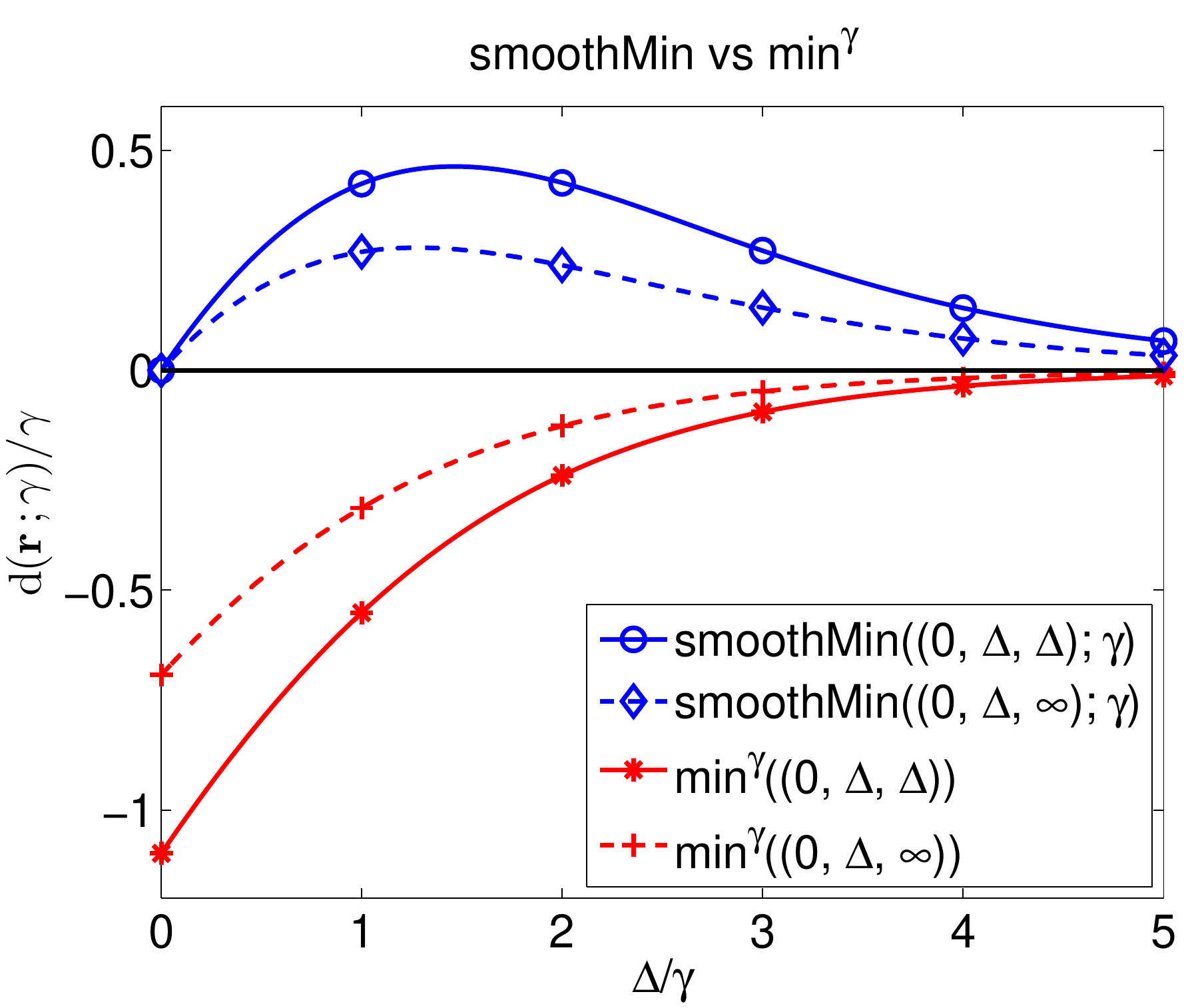}%
  \caption{The penalty terms $d(\mathbf{r}; \gamma)$ $=$ $s(\mathbf{r}; \gamma) - \min(\mathbf{r})$ for the two smooth min approximations given in (\ref{eq:smoothmin}) and (\ref{eq:min_gamma}). For simplicity we can assume the $r_i$'s are in sorted order.  We evaluate two extreme cases where $r_2 = r_1 + \Delta$ for some $\Delta \geq 0$ and the next largest value, $r_3$, is either equal to $r_2$ or much larger.  It can be shown that $d(\mathbf{r}; \gamma) = \gamma d(\mathbf{r}/ \gamma; 1)$ and therefore we need only plot one scale-invariant curve for each case.} %
\label{fig:relaxed_min_comparison}
\end{figure}

While both the $\text{min}^\gamma$ and $\text{smoothMin}$ operators are differentiable approximations of the min operator (with the min operator subsumed as a special case), their different behaviours have profound effects on learning in our setting.  These differences are illustrated in the plot in Fig.\ \ref{fig:relaxed_min_comparison}, where 
without loss of generality we assume the costs 
are sorted in increasing order, with $r_2 = r_1+\Delta$ for some $\Delta \geq 0$ and $r_3 \geq r_2$.
As can be seen, the $\text{min}^\gamma$ function is strictly monotonically increasing.  As a result, with all other things being equal, minimizing this function encourages ties, \ie, the penalty is minimized only when $\Delta = 0$. This is an undesirable behaviour 
as we seek the resulting embeddings to yield a well-defined path (\ie, optimal alignment) in our cumulative cost matrix, $\mathbf{R}(i,j)$.  %
In contrast, our $\text{smoothMin}$ operator defines a contrastive watershed (at approximately $1.5\gamma$), 
where values to the left of the watershed encourage ties, while to the right the values are encouraged to be well separated.  Moreover, since $d(\mathbf{r}; \gamma) \geq 0$ for $\text{smoothMin}$, our smoothDTW approach always provides an upper bound on the cost of the optimal path. The supplemental presents an expanded discussion and comparison. %

\subsection{Contrastive cost function} \label{sec:contrastive_cost}

To complete the definition of our smoothDTW recurrence in (\ref{eq:softDTW}),  
we now specify the cost function $c(\mathbf{x}_i, \mathbf{y}_j)$.  Specifically, for each element $\mathbf{x}_i$ in $\mathbf{X}$, we wish to express the cost of matching $\mathbf{x}_i$ to any single item $\mathbf{y}_j$ in $\mathbf{Y}$, given that at least one of the elements in $\mathbf{Y}$ must match. %
Moreover, in keeping with our probabilistic path finding formulation, $c(\mathbf{x}_i, \mathbf{y}_j)$ should be the negative log probability of matching the given $\mathbf{x}_i$ to a selected $\mathbf{y}_j$ in $\mathbf{Y}$. This leads to the (non-symmetric) contrastive formulation
 \begin{align}
   c(i, j; \mathbf{X},\mathbf{Y}) &= -\log(\text{softmax}_2(\tilde{\mathbf{X}}^\top\tilde{\mathbf{Y}}; \beta))_{i,j} ,\\
      &= -\log\left(\frac{\exp(\tilde{\mathbf{x}}_i^\top \tilde{\mathbf{y}}_j/\beta)}{\sum_{k=1}^N \exp(\tilde{\mathbf{x}}_i^\top \tilde{\mathbf{y}}_k/\beta)}\right), \label{eq:contrastive_loss}
\end{align}
where the $\text{softmax}_2$ operator is defined as the standard $\text{softmax}$ with a temperature hyper-parameter, $\beta$, over the second matrix dimension, \ie columns. Also the notation $\tilde{\mathbf{x}}_i$ denotes $L_2$ normalization, so $\tilde{\mathbf{x}}_i$ $=$  $\mathbf{x}_i/||\mathbf{x}_i ||_2$, and so on.
The use of the negative log softmax operator 
over the set of correspondences in (\ref{eq:contrastive_loss}) encourages a (soft) winner-take-all, where one pair $(\mathbf{x}_i, \mathbf{y}_j)$ has a significantly higher cosine similarity than all the other options.  The normalization across the column penalizes situations without a clear winner.

With this cost function the resulting values $\mathbf{R}(i, j)$ are the optimum value of the negative log probability for any feasible path starting at $(0,0)$ and ending at $(i,j)$.  Here, this negative log probability is the sum of the matching cost, $c(\mathbf{x}_m,\mathbf{y}_n)$ (cf.\ (\ref{eq:contrastive_loss})), and the smoothness penalty 
$d(\mathbf{r}_{m,n})$ (cf.\ (\ref{eq:softDTWstep1})), at each vertex, $(m, n)$, along the optimal path ending at $(i, j)$. Correspondingly, we define the alignment loss for matching $\mathbf{X}$ to $\mathbf{Y}$ as:
\begin{equation}
   \mathcal{L}_\text{smoothDTW}(\mathbf{X},\mathbf{Y}) = \mathbf{R}(M, N). \label{eq:align_loss}
\end{equation}
We use the sum of the alignment losses for matching $\mathbf{X}$ to $\mathbf{Y}$ and vice versa as the overall alignment loss, as shown in the left panel in Fig.\ \ref{fig:loss}.

The combination of the softmax over elements in $\mathbf{Y}$, in (\ref{eq:contrastive_loss}), and the use of smooth DTW to formulate the alignment cost (\ref{eq:align_loss}), rewards embeddings that are both: a) contrastive across the elements in $\mathbf{Y}$; and, moreover, b) have their best matching pairs $(\mathbf{x}_i, \mathbf{y}_j)$ arranged along a feasible path from $(0, 0)$ to $(M, N)$. We show in our ablation study that the ability to leverage these two properties during training are  key to our utilization of the temporal alignment proxy.  In contrast, dropping the softmax and simply computing the inner-product between elements could lead to the collapse of the embeddings around a single point during training, thus allowing the network to trivially minimize the alignment cost. 
To avoid such collapse, previous DTW-based methods have resorted to  adding a discriminative loss downstream \cite{ChangHS0N19,CaoJCCN20}.

\subsection{Global cycle-consistency} \label{sec:cyclic}

An additional loss is based on the notion that the match from sequence $X$ to $Y$, composed with the match from $Y$ to $X$, should ideally be the identity.  We  formulate this directly in terms of the cumulative cost matrix $\mathbf{R}_{X,Y}$ for matching sequence $X$ to $Y$ (as defined by (\ref{eq:softDTW}), (\ref{eq:smoothmin}), and (\ref{eq:contrastive_loss})), along with the cost matrix for matching $Y$ to $X$, namely $\mathbf{R} _{\,Y,X}$.  Given the interpretation that $\mathbf{R}_{X,Y}(i, j)$ is the optimal negative log probability of a path from $(0,0)$ to $(i,j)$ for matching $X$ to $Y$, consider the implied conditional distribution for matching the  prefix sequences $X(1\!:\!i)$ to $Y(1\!:\!j)$ for different $j$'s, namely
\begin{align}
    p_{X,Y}(j \,|\, i) &:= [\text{softmax}_2(-\mathbf{R}_{X,Y}/\alpha)]_{i,j} \nonumber \\
                   &= \frac{e^{-\mathbf{R}_{X,Y}(i, j)/\alpha}}{\sum_{k=1}^N e^{-\mathbf{R}_{X,Y}(i, k)/\alpha}}.
                   \label{eq:pXY}
\end{align}
Note that this distribution does not use any information for elements $k > i$ from sequence $X$, and is only obtained from the forward pass of matching $X$ to $Y$. We use the notation 
\begin{equation}
    P_{X,Y} := [\text{softmax}_2(-\mathbf{R}_{X,Y}/\alpha)]^\top 
                   \label{eq:PXY}
\end{equation}
to denote the $N \times M$ matrix with elements $(\mathbf{P}_{X,Y})_{n,m}$ $=$ $p_{X,Y}(n \,|\, m)$.

Ideally, the %
contrastive matching in (\ref{eq:contrastive_loss}) is sharp and forms a feasible path from $(0,0)$ to $(i,j)$, thereby providing a strongly peaked conditional distribution $p_{X,Y}(j | i)$ for each $i$.  %
However, without knowing the ground truth matching $\{(i_k, j_k)\}_{k=0}^L$, we cannot use an explicit log-likelihood loss.  This issue can be avoided by considering the composed conditional distribution
\begin{equation}
 p_{X,Y,X}(j \,|\, i) := \sum_{k=1}^N p_{\,Y,X}(j \,|\, k) \, p_{X,Y}(k \,|\, i),
 \label{eq:back_and_forth}
\end{equation}
which is formed by treating $p_{X,Y}(k \,|\, i)$ and $p _{\,Y,X}(j \,|\, k)$ as conditionally independent distributions.  It is easy to verify that
this is indeed a distribution over elements $j$ of $X$.  Moreover, following the above matrix notation, it is represented by the $M \times M$ matrix $\mathbf{P}_{Y,X} \mathbf{P}_{X,Y}$.

In the ideal case, this transport from elements in one sequence to another 
and back again should return to the same starting element.  From (\ref{eq:back_and_forth}) this corresponds to $\mathbf{P}_{Y,X} \mathbf{P}_{X,Y}$ equaling the 
identity matrix, $\mathbf{I}_{M\times M}$.
Thus, our global cycle-consistency loss is the sum of cross-entropy losses:
\begin{equation}
       \mathcal{L}_\text{GCC}(\mathbf{X},\mathbf{Y}) = -\sum_{i=1}^M\log( (\mathbf{P}_{Y,X} \mathbf{P}_{X,Y})_{i,i}).\label{eq:gcc}
\end{equation}
The right panel in Fig.\ \ref{fig:loss} provides a summary of our global cycle-consistency loss.

\subsection{Training and implementation details} \label{sec:training details}
Our final loss function is obtained by combining the contrastive alignment loss, (\ref{eq:align_loss}), and the global cycle consistency loss, (\ref{eq:gcc}), according to
\begin{align}
\mathcal{L}(&\mathbf{X},\mathbf{Y}) = \lambda_g\mathcal{L}_\text{GCC}(\mathbf{X},\mathbf{Y}) \nonumber\\
&+\lambda_s(\mathcal{L}_\text{smoothDTW}(\mathbf{X},\mathbf{Y}) +  \mathcal{L}_\text{smoothDTW}(\mathbf{Y},\mathbf{X})), \label{eq:softDTWstep1} 
\end{align}
where $\lambda_g$ and $\lambda_s$ are weights used to balance the two losses and are empirically set to $1.0$ and $0.1$, respectively. The temperature hyper-parameters, $\gamma$ and $\beta$ used in (\ref{eq:smoothmin}) and (\ref{eq:contrastive_loss}) are both set to $0.1$, while $\alpha$ in (\ref{eq:pXY}) is $1$.

This overall loss is used to train a convolutional architecture composed of a backbone encoder applied framewise followed by an embedding network. Specifically, ResNet50-v2 \cite{HeZRS16} is used as our backbone encoder where we extract features from {\tt Conv4c} layer.  We adopt the same embedding network used in previous related work \cite{DwibediATSZ19} comprised of two 3D convolutional layers, a global 3D max pooling layer, two fully connected layers, and a linear projection layer.  The final embedding is L2 normalized. To learn over sequence pairs $(\mathbf{X},\mathbf{Y})$, we randomly extract $T=20$ frames from each sequence. Sampling of video frames is random to avoid learning potential trivial solutions that may arise from strided sampling. For a fair comparison to previous approaches using the same architecture (\ie, \cite{DwibediATSZ19, SermanetLCHJSLB18, MisraZH16}), we use the same batch size of four sequences.  Finally, our learning rate is fixed to $10^{-4}$ for all our experiments.

\section{Empirical evaluation}
We evaluate the efficacy of our learned embeddings on challenging temporal fine-grained tasks, thereby going beyond traditional clip-level recognition tasks. In particular, our proposed loss is evaluated on fine-grained action recognition (\ie, action phase classification), few-shot fine-grained classification, and video synchronization. %
In addition, we also show that learning to align temporal sequences supports different downstream applications such as synchronous playback, 3D pose reconstruction and fine-grained audio/visual retrieval.

\subsection{Datasets}
To evaluate our method, we use
the PennAction \cite{PennAction}
and FineGym \cite{shao2020finegym} video datasets. Both datasets contain a diverse set of videos
of human-related sports or fitness activities. 
These datasets are selected as they allow for learning framewise alignments and evaluating on tasks
where fine-grained temporal distinctions are critical.

\noindent{\bf PennAction} \cite{PennAction} contains 2326 videos of humans performing 15 different sports or fitness actions. The videos are tightly cropped temporally around the start and end of the action. Notably, 13 %
categories contain non-repetitive actions.
The dataset also includes ground truth 2D keypoint labels which we later use to demonstrate a 3D pose reconstruction application grounded on our alignment method.

\noindent{\bf FineGym} \cite{shao2020finegym} is a recent large-scale fine-grained action recognition dataset that was specifically designed to evaluate the ability of an algorithm to parse and recognize the different phases of an action. Each video in FineGym is annotated according to a  three-level hierarchy denoting the \textit{event} being performed in the video, the different \textit{sets} involved in performing the event, and the framewise \textit{elements} (\ie, action phases) involved in each set. To perform any \textit{event-level} action, a gymnast may perform the different sets in any order. To train embeddings using our alignment-based method, we re-organize the FineGym dataset such that all sets belonging to the same event appear in the same order in any given video. An example of this re-organization is provided in the supplemental. Also, it should be noted that while the videos of the vault event (VT) in FineGym depict the gymnast performing the action in three phases, the first two phases of the action are not explicitly labeled in the original dataset. For the sake of completeness we use the provided start and end times for these phases, thereby adding two new fine-grained action labels in FineGym. The remaining events in FineGym  are otherwise unchanged. To account for these minor additions to the annotations, we refer to the extensions of FineGym99 and 288 as FineGym101 and FineGym290, respectively. Notably, we also report results on the original FineGym99 and 288 in the supplemental.
Importantly, we do not use the \textit{element-level} (\ie, action phase) labels during training. 

\subsection{Baselines}
We compare our approach to other weakly supervised \cite{DwibediATSZ19,SermanetLCHJSLB18,ChangHS0N19} and self-supervised \cite{BenaimELMFRID20,MisraZH16} methods that entail temporal understanding in their definition. A detailed description of the baselines is provided in the supplemental.

\subsection{Ablation study}
We first present an ablation study that validates the contribution of each component of our loss.
For this purpose, we evaluate fine-grained action recognition performance on FineGym101. Following previous work \cite{DwibediATSZ19}, we use a Support Vector Machine (SVM) classifier \cite{cortes1995support} on top of the learned embeddings to report framewise fine-grained
classification accuracy. Notably, the classifier is trained on the extracted embeddings with no additional fine-tuning of the network. 
The results in Table \ref{tab:ablation} show the pivotal role of our contrastive cost. In fact, turning off the contrastive component of our cost, (\ref{eq:contrastive_loss}), and simply relying on the cosine distance always leads to no improvement in learning from the onset of training.
Also, these results show the advantage of adding our global cycle consistency, which further validates the correspondences. Finally, we also compare the performance of our $\text{smoothMin}$ definition vs.\ the more widely used $\text{min}^\gamma$. The superiority of the adopted smooth definition supports the laid out arguments in Section \ref{sec:smooth-min}.
\vspace{-5pt}
\begin{table}
	\begin{center}
		\resizebox{0.9\columnwidth}{!}{
			\begin{tabular}{c|c|c|c}
				\cline{2-4}
				&  Contrastive-Cost & GCC & FineGym101 \\
				\hline
				\multirow{4}{*}{SmoothDTW} & - & - & 28.20\\
				& - & \checkmark & 28.20 \\
				& \checkmark & - & 47.32\\
				& \checkmark & \checkmark & \textbf{49.51}\\
				\hline
				$\text{min}^\gamma$ & \checkmark & \checkmark & 48.07 \\
				\hline
			\end{tabular}
		}
	\end{center}
	\caption{Ablation study of the various components of our loss function, (\ref{eq:softDTWstep1}).
	 Contrastive-Cost refers to our columnwise contrastive cost, (\ref{eq:contrastive_loss}), and GCC refers to our global cycle consistency loss, (\ref{eq:gcc}).}
	\label{tab:ablation}
	\vspace{-15pt}
\end{table}

\subsection{Fine-grained action recognition}
We now compare our fine-grained action recognition performance to our baselines using the FineGym dataset and consider two training settings for the backbone framewise encoder. \textbf{(i) scratch:} the backbone ResNet50 \cite{HeZRS16} is trained from scratch with our proposed loss,  \textbf{(ii) only-bn:} we fine-tune batch norm layers of ResNet50 from a model pre-trained on ImageNet \cite{ILSVRC15}. The embedder is otherwise trained from scratch in both cases. A third setting where all layers are fine-tuned is presented in the supplemental.

The results summarized in Table \ref{tab:SOTA-comparisoln} (and the supplemental) speak decisively in favour of our method, where we outperform all other weakly and self-supervised methods with sizable margins. The gap is especially striking in the case of
SpeedNet. %
This poor performance can largely be attributed to the fact that the task optimized in SpeedNet does not require detailed framewise understanding. On the other hand, the closest approach to ours is TCC as it also relies on pairwise local matchings between videos of the same class; however, the matchings
are realized independently and thus ignores informative long-term sequence structure.
This is in contrast to SaL which uses frames from the same videos to solve tasks requiring temporal understanding. Importantly, the superiority of our results compared to TCC
demonstrates that the global nature of our loss makes the learned embeddings more robust to the presence of repeated sub-actions as is the case for most of the FineGym videos. Interestingly, the results obtained with D$^3$TW* highlight the limits of the downstream discriminative loss, which requires an explicit construction of positive and negative examples for training.
Notably, while our method outperforms all alternatives under both training settings, the best overall results are obtained under the \textit{only-bn} setting and it is therefore used for  
all other experiments reported in this paper.

We also considered classification results of each event separately where we also outperformed all alternatives. Importantly, visualizations of the learned features suggests that the proposed loss learns to adapt and identify the most reliable cues to learn the alignments. Please see supplemental for detailed results, discussions, and visualizations.

\begin{table}[t]
	\begin{center}
	\resizebox{0.9\columnwidth}{!}{
		\begin{tabular}{c|c|c|c}
			\hline
			Method & Training & FineGym101 & FineGym290 \\
			\hline
			SpeedNet \cite{BenaimELMFRID20} & \multirow{6}{*}{scratch} & 30.40& 29.87\\
			TCN \cite{SermanetLCHJSLB18} & & 36.52 & 37.40 \\
			SaL \cite{MisraZH16} & & 40.25 & 37.98 \\
			D$^3$TW* \cite{ChangHS0N19} & & 32.10 & 32.15\\
			TCC \cite{DwibediATSZ19} &  & 41.78 & 40.57 \\
			Ours &  &\textbf{45.79} & \textbf{43.49} \\
			\hline
			SpeedNet \cite{BenaimELMFRID20} &  \multirow{6}{*}{only bn} & 34.38 & 35.92 \\
			TCN \cite{SermanetLCHJSLB18} & & 41.75 & 39.93 \\
			SaL \cite{MisraZH16} && 42.68 & 41.58 \\
			D$^3$TW* \cite{ChangHS0N19} &  & 38.21 & 34.04 \\
			TCC \cite{DwibediATSZ19} &  & 45.62& 43.40 \\
			Ours & &\textbf{49.51} & \textbf{46.54} \\
			\hline
			
		\end{tabular}
	}
	\end{center}
	\caption{Fine-grained action recognition accuracy on both organizations of FineGym.}
	\label{tab:SOTA-comparisoln}
	\vspace{-15pt}
\end{table}

\subsection{Few-shot fine-grained action recognition}
An advantage of our proposed weakly supervised method is that it does not rely on framewise labels for training. To evaluate this advantage, we also report few-shot classification results. In this case, the entire training set is used to learn the embeddings, but only a few videos per class are used to train the classifier. In particular, we use FineGym101 for this experiment with an increasing number of videos per class  to train the classifier, starting from the 1-shot setting all the way to using the entire dataset.

\begin{figure}[t]
\centering
	\resizebox{0.7\columnwidth}{!}{
	\includegraphics{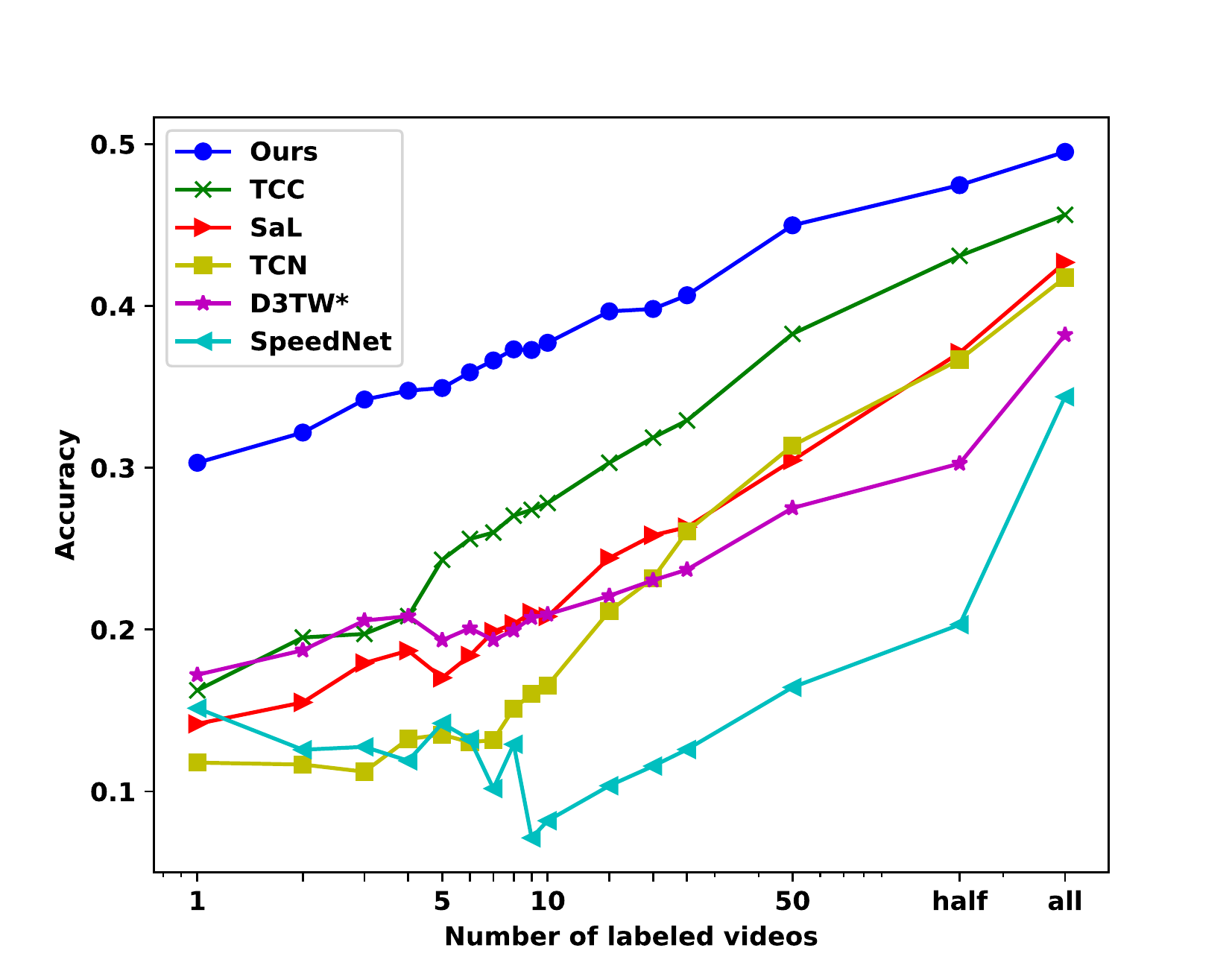}
    }
	\caption{Few-shot fine-grained action recognition accuracy on FineGym101.}%
	\label{fig:few-shot}
	\vspace{-5pt}
\end{figure}

The plot in Fig.\ \ref{fig:few-shot} further confirms the superiority of our proposed method.  As can be seen, our method outperforms all others across the range of number of labeled training videos used, with an especially strong performance even under the challenging 1-shot setting.
\subsection{Video synchronization}
To evaluate the quality of the synchronization (\ie, alignment) between two videos we use  Kendall's Tau metric, which does not require framewise labels.  However, this metric assumes little to no repetitions in the aligned videos. We therefore follow previous work \cite{DwibediATSZ19} and report results only on the 13 classes without repetition in PennAction %
(\ie,  strumming guitar and jumping rope are not included). Importantly, while previously reported results \cite{DwibediATSZ19} were obtained by training a different network for each class of PennAction, we consider the more challenging setting of learning a joint representation over all classes (\ie, as done in all previous experiments with FineGym). For a fair comparison, all baselines are also trained over all classes of PennAction rather than one network per class.
\begin{table}[t]
	\begin{center}
		\resizebox{0.85\columnwidth}{!}{
		\begin{tabular}{c|c|c}
			\hline
			\multirow{1}{*}{Method} & \multicolumn{1}{c|}{Kendall's Tau} & \multicolumn{1}{c}{Phase classification}\\
			\hline
			TCN \cite{SermanetLCHJSLB18} & 65.29 & 69.3 \\
			SaL \cite{MisraZH16} & 53.87 & 69.4 \\
			TCC \cite{DwibediATSZ19} & 71.0  & 77.51\\
			\hline
			Ours & \textbf{74.84} & \textbf{78.90} \\
			\hline
			
		\end{tabular}
	}
	\end{center}
	\caption{Video alignment results using Kendall's Tau metric and action phases classification on PennAction. Different from previous work, these results were obtained by training a single network for all classes, thereby learning a joint representation.}%
	\label{tab:Kendall}
	\vspace{-15pt}
\end{table}

Table \ref{tab:Kendall} summarizes our video alignment results, where we once again achieve state-of-the-art performance compared to the baselines. Importantly, our performance is also superior to previously reported results under the multi-network training setting \cite{DwibediATSZ19}. We report results under this setting in the supplemental.

In addition to the quality of synchronization, we also report in Table \ref{tab:Kendall} state-of-the-art results on action phase classification. Notably, the action phase labels used here are provided by the original authors of the PennAction dataset \cite{PennAction} as the labels used previously \cite{DwibediATSZ19}  were not made publically available.

\vspace{-5pt}
\section{Downstream applications}
\begin{figure}
	\begin{center}	
		\resizebox{0.8\columnwidth}{!}{
			\begin{tabular}{c}
			\includegraphics{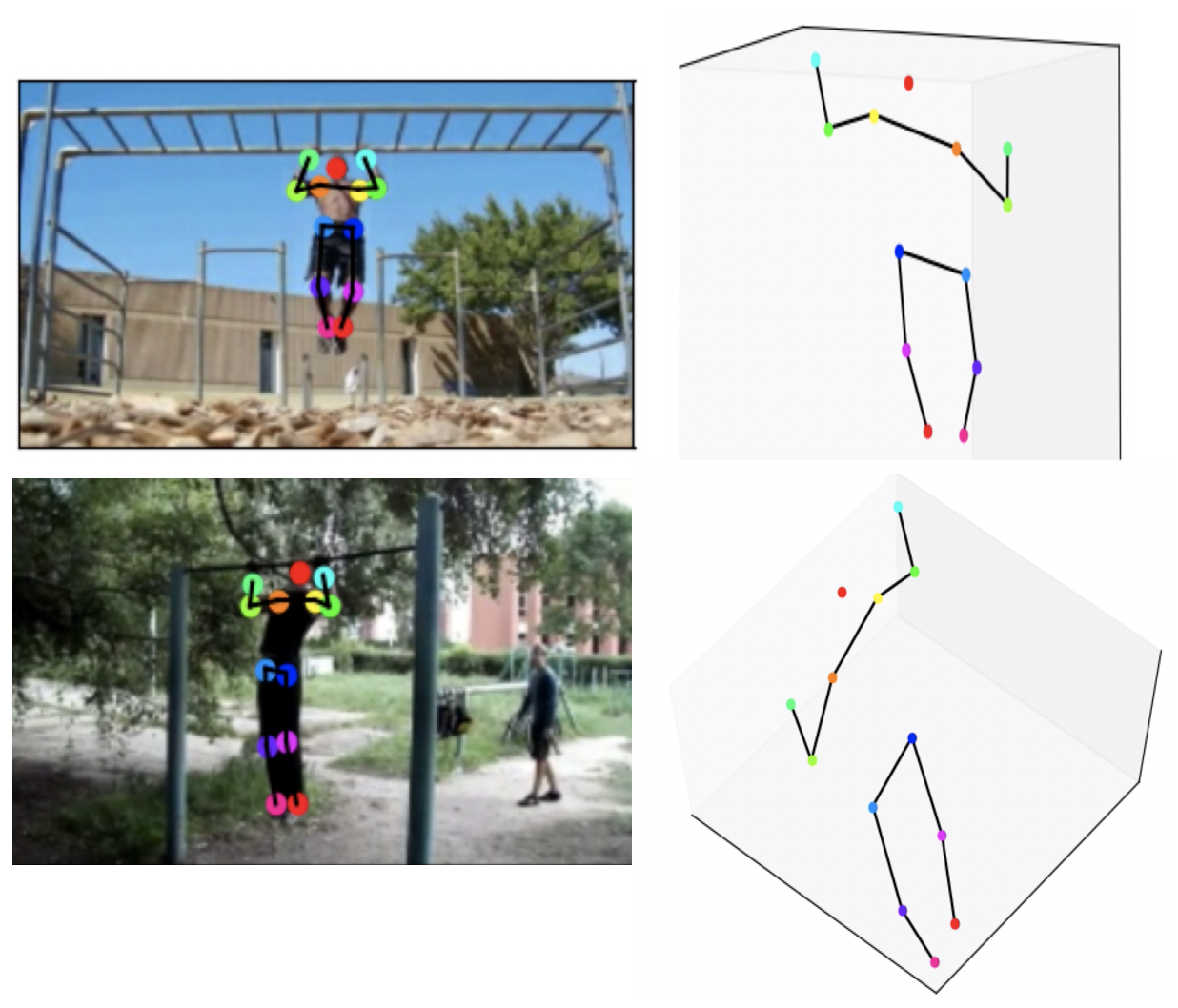} \\
            \end{tabular}	
	}
	\end{center}
	\vspace{-5pt}
\caption{3D pose reconstruction results derived from the learned alignments. Synchronized video frames from two sample training videos in the pullup category of PennAction with overlayed 2D keypoints are shown on the left. Reconstructed 3D pose from two different viewpoints is shown on the right. %
}
\label{fig:3D_POSE}
\vspace{-15pt}
\end{figure}

\noindent{\bf 3D pose reconstruction.} As demonstrated by our evaluations, our proposed weakly-supervised method is capable of temporally aligning videos of similar actions even while they are captured in different environments, at different execution rates, and from different viewpoints. As a result, we can align poses of the same action from different viewpoints by forcing different videos to play synchronously. Examples of this video synchronization are provided in the supplemental. Importantly, the synchronization of videos taken at various viewpoints can serve as the basis for 3D pose reconstruction. To demonstrate this ability, we use videos from PennAction and their corresponding ground truth 2D keypoint labels. In particular, given a random query video from a given class in PennAction, we first start by aligning the remaining videos (from the same class) to it. Given these aligned frames and their corresponding 2D keypoints we use the Tomasi-Kanade factorization algorithm \cite{TomasiK92}, followed by bundle adjustment \cite{Hartley-Zisserman-Book}, to compute a temporally aligned 3D model of the action performed.
Fig.\ \ref{fig:3D_POSE} presents an example of our 3D pose reconstructions; 
additional results are available in the supplemental material.

\noindent{\bf Audio-visual alignment.} Finally, we demonstrate that our alignment based representation learning method can be applied to other types of sequences by applying it on separate audio and visual inputs. Given that any audio-visual pair is by default aligned and to avoid learning trivial solutions, we sample audio segments and video frames differently to make the task of learning the alignment harder on the networks and consequently learn strong audio and visual embeddings. In particular, while the audio signal is uniformly sampled into consecutive one second long segments, video frames are on the other hand randomly sampled along the temporal dimension. This sampling strategy encourages our model to learn different alignment paths due to the randomness in the video frame selection. For visual features, the same backbone and embedding network described  in Sec.\  \ref{sec:training details} is used to encode video frames, while we use VGGish \cite{VGGish} to encode audio signals. In particular, each one second long audio segment is first converted into a log mel spectogram and used as an input for the VGGish network. Training is otherwise performed as described in Sec.\ \ref{sec:training details}.

For this sample application, we use the firing cannon class from the VGGSound dataset \cite{VGGSound} for training and testing. This class is selected as it is %
strongly visually indicated %
with an easily identifiable salient auditory signal (\ie, the explosion sound emitted upon firing of a cannon).

\begin{figure}[t]
	\begin{center}	
		\resizebox{0.99\columnwidth}{!}{
				\includegraphics{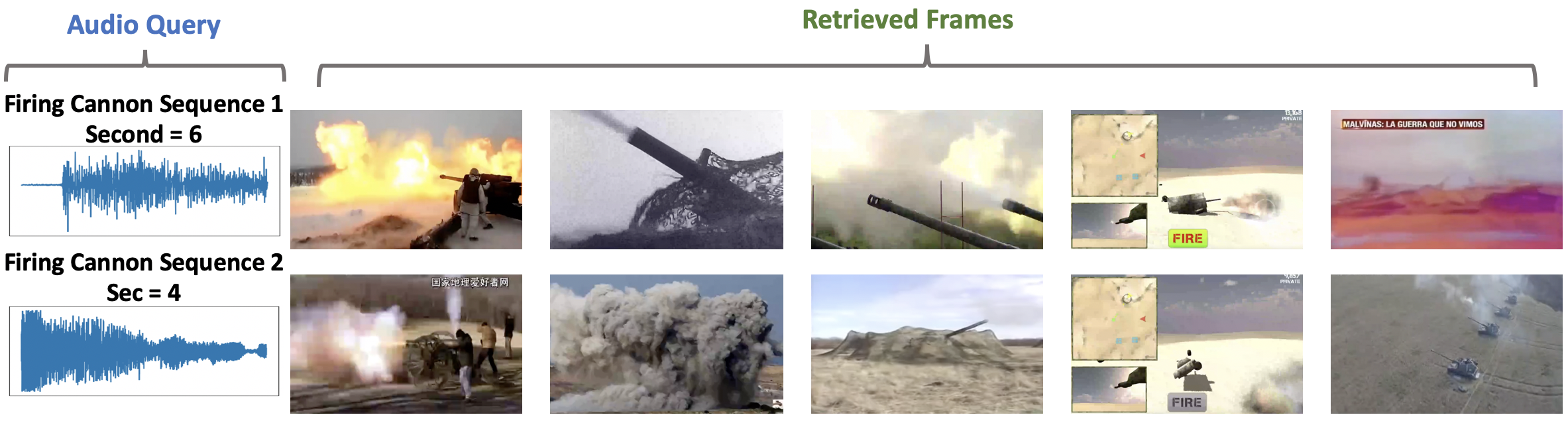}
		}
	\end{center}
	\vspace{-5pt}
	\caption{Sample fine-grained audio/visual retrieval using nearest neighbor matches in embedding space, showing that the audio of the explosion can be used to retrieve corresponding visual event.}
	\label{fig:aud2vid}
	\vspace{-15pt}
\end{figure}

To demonstrate the quality of the learned audio and visual embeddings, we evaluate them on the task of fine-grained audio/visual retrieval. In particular, given a one second long audio signal corresponding to the moment of firing a cannon, we extract its corresponding audio embedding from the VGGish network trained using our approach. This embedding is then used to query corresponding visual embeddings from the vision network. The top-5 nearest frame embeddings from all videos in the test set are extracted. Sample correspondences, shown in Fig.\ \ref{fig:aud2vid}, clearly depict that the corresponding visual embeddings also capture the moment of the firing visually. More details and sample results are presented in the supplemental. %

\section{Conclusion}
In summary, this work introduced a novel weakly
supervised method for representation learning relying on sequence alignment as a supervisory signal and taking a probabilistic view in tackling this problem. Because the latent supervisory signal entails detailed temporal understanding, we judge the effectiveness of our learned representation on tasks requiring fine-grained temporal distinctions and show that we establish a new state of the art. In addition, we present two applications
of our temporal alignment framework, thereby opening up new avenues for future investigations grounded on the proposed approach.

{\small
\bibliographystyle{ieee_fullname}
\bibliography{bibref_short,bibref}
}

\clearpage
\appendix
\section{Supplementary material}

Our supplementary material is organized as follows: Sec.\ \ref{sec:properties}
derives several properties of the two smooth minimum approximations introduced 
in the main manuscript. Sec.\ \ref{sec:baselines} provides details of our evaluation baselines. 
Sec.\ \ref{sec:FGR} further explains our re-organization of FineGym to satisfy our training requirements. For completeness, Secs.\ \ref{sec:FGR1} and \ref{sec:FGR2} provide additional fine-grained action recognition results. Similarly, Sec.\ \ref{sec:PennR} provides additional alignment results on PennAction. Finally, Sec.\ \ref{sec:visualization} probes our learned representations via visualizations to ascertain what information is captured. A supplemental video highlighting the different applications of the proposed loss is also provided.

\section{Smooth Minimum Properties} \label{sec:properties}

We derive several properties of the two smooth minimum approximations, smoothMin and $\text{min}^\gamma$. For more general applicability, we consider any vector-valued input, $\mathbf{a} \in \mathbb{R}^N$. \\

\noindent{\bf smoothMin properties.} The smoothMin function is defined by
\begin{align}
\text{smoothMin}(\mathbf{a}; \gamma) = 
\left\{
\begin{matrix}
   \min\{a_i ~|~ 1 \leq i \leq N \},& \gamma = 0\\
   \frac{\sum_{i=1}^N a_i e^{-a_i/\gamma}} {\sum_{j=1}^N e^{-a_j/\gamma}},  &  \gamma > 0
\end{matrix}\right., \label{eqApp:smoothmin}
\end{align}
where $\gamma$ denotes a temperature hyper-parameter.  Define $a_0$ to be the minimum coefficient in $\mathbf{a}$, that is, $a_0 = \min( \mathbf{a} )$.  Then for $\gamma > 0$ it follows that
\begin{align}
\text{smooth}&\text{Min}(\mathbf{a}; \gamma) = \frac{\sum_{i=1}^N a_i  e^{-(a_i-a_0)/\gamma}} {\sum_{j=1}^N e^{-(a_j - a_0)/\gamma}} \nonumber \\
&= a_0 + \frac{\sum_{i=1}^N (a_i- a_0)  e^{-(a_i-a_0)/\gamma}} {\sum_{j=1}^N e^{-(a_j - a_0)/\gamma}} \nonumber \\
&= a_0 + \gamma \frac{\sum_{i=1}^N ( (a_i - a_0)/\gamma ) e^{-(a_i-a_0)/\gamma}} {\sum_{j=1}^N e^{-(a_j - a_0)/\gamma}}  \nonumber\\
&= \min( a ) + \gamma \; \text{smoothMin}((\mathbf{a} - a_0) / \gamma; 1). 
\end{align}
Therefore, the smoothness penalty satisfies
\begin{align}
d( \mathbf{a}; \gamma) &:= \text{smoothMin}(\mathbf{a}; \gamma) - \min(\mathbf{a}),\nonumber \\
 &= \gamma \; \text{smoothMin}((\mathbf{a} - a_0) / \gamma; 1). \label{eqApp:penaltySmoothMin}
\end{align} 
Note that from (\ref{eqApp:smoothmin}) it follows that $d(\mathbf{a} ; \gamma ) \geq 0$, so we have $\text{smoothMin}(\mathbf{a}; \gamma) \geq \min( \mathbf{a} )$, \ie, it provides an upper bound on the true minimum.

The maximum possible value of the penalty $d( \mathbf{a}; \gamma)$ $=$ $\gamma \text{smoothMin}(\mathbf{a} - a_0) / \gamma; 1)$ is also of interest.  Given (\ref{eqApp:penaltySmoothMin}) it is sufficient to maximize $\text{smoothMin}(\mathbf{r}; 1)$ subject to $r_1 =0$ and $r_i \geq 0$ for $i \in \{2, \ldots N\}$.  By setting $\frac{\partial}{\partial r_i} \text{smoothMin}(\mathbf{r}; 1) = 0$ and simplifying, we find
\begin{equation}
    1 - r_i + \text{smoothMin}(\mathbf{r}; 1) = 0 \label{eqApp:maxPenalty_ri}
\end{equation}
for $i\geq 2$.  This implies all the $r_i$'s for $i>1$ are equal at the maximum.  Using (\ref{eqApp:smoothmin}) in (\ref{eqApp:maxPenalty_ri}), setting $r_1 = 0$ and $r_i = x$ for $i>1$, and simplifying, we find $x$ must be the solution of
\begin{equation}
    x - 1 = (N-1) e^{-x}. \label{eqApp:maxPenalty_ri2}
\end{equation}
For $N \geq 1$, (\ref{eqApp:maxPenalty_ri2}) has a unique solution $x(N) \geq 1$.
This implies that the penalty function (\ref{eqApp:penaltySmoothMin}) only has maxima, $\mathbf{a}$, for which there is exactly one distinct minimum value $a_j = \min(\mathbf{a})$, and all the other values are in an $(N-1)$-way tie for second largest at $a_i = a_j + x \gamma$ for $i \neq j$, and where $x$ is as in (\ref{eqApp:maxPenalty_ri2}). 

To compute the value of this maximum, we substitute the resulting vector $\mathbf{r} = \begin{bmatrix} 0 & x(N)& \ldots& x(N)\end{bmatrix}^\top$ 
into $\text{smoothMin}(\mathbf{r}; 1)$ and, by simplifying, we find
\begin{align}
    \max d(\mathbf{r}; 1) &= \frac{(N-1) x(N) e^{-x(N)}}{1+(N-1)e^{-x(N)}} \nonumber \\
    &= \frac{(N-1) x(N) \left[(x(N)-1)/(N-1)\right]}{1+(N-1)\left[(x(N)-1)/(N-1)\right]} \nonumber \\
    &= x(N) - 1, \label{eqApp:maxPenalty}
\end{align}
where we used (\ref{eqApp:maxPenalty_ri2}) in the second line above. 

For concrete examples, we find $x(2) \approx 1.2785$, $x(3) \approx 1.4631$ for $N = 2$ and $3$, respectively.  Also, for $N \geq 4$ it follows from (\ref{eqApp:maxPenalty_ri2}) that $x(N) < \log(N+1)$ (since the left hand side less minus the right is negative at $x=0$, positive for $x = \log(N+1)$, and the derivative of this difference with respect to $x$ is positive).  Therefore, (\ref{eqApp:maxPenalty}) implies that the maximum smoothness penalty when $N = 2$ is roughly $0.2785 \gamma$, and $0.4631 \gamma$ for $N = 3$, which are in agreement with the plot in %
Fig.\ 3 of the main manuscript. Moreover, for large $N$ the maximum penalty is $O(\gamma \log(N))$.\\

\noindent{\bf $\text{min}^\gamma$ properties.} Similarly we examine $\text{min}^\gamma$, which is defined by
\begin{align}
\text{min}^\gamma(\mathbf{a}) = 
\left\{
\begin{matrix}
   \min\{a_i\}, & \gamma = 0 \\
   -\gamma \log \sum_{i=1}^N e^{-a_i/\gamma} ,  &  \gamma > 0
\end{matrix}\right.. \label{eqApp:min_gamma}
\end{align}
For $a_0 = \min(\mathbf{a})$ and $\gamma > 0$ we have
\begin{align}
\text{min}^\gamma(\mathbf{a}) &= -\gamma \log \sum_{i=1}^N \left[ e^{-(a_i-a_0)/\gamma} e^{-a_0/\gamma}\right],
 \nonumber \\
  &= -\gamma \log \left[ e^{-a_0/\gamma} \sum_{i=1}^N  e^{-(a_i-a_0)/\gamma} \right],
 \nonumber \\
  &= -\gamma \left[ -a_0/\gamma + \log \sum_{i=1}^N e^{-(a_i-a_0)/\gamma} \right], \nonumber \\
 &= \min(\mathbf{a}) - \gamma  \log \sum_{i=1}^N e^{-(a_i-a_0)/\gamma}. \label{eqApp:minGammaDiff}
\end{align}
Recall that $a_0 = \min(\mathbf{a})$ so $a_i = a_0$ for at least one $i$.  Given that all the terms in the sum $\sum_{i=1}^N e^{-(a_i-a_0)/\gamma}$ are non-negative, and one of them is 1, we conclude that $\sum_{i=1}^N e^{-(a_i-a_0)/\gamma} > 1$.  Therefore, we have $\text{min}^\gamma (\mathbf{a}) < \min( \mathbf{a} )$, \ie, it provides an lower bound on the true minimum.

Moreover, the maximum sum (\ref{eqApp:minGammaDiff}) occurs when $a_i = a_0$ for all $i$ (\ie, there is an $N$-way tie for the minimum). In this case $\sum_{i=1}^N e^{-(a_i-a_0)/\gamma} = N$.  Therefore, we have 
\begin{equation}
    \sum_{i=1}^N e^{-(a_i-a_0)/\gamma} \in (1, N].
\end{equation}
Finally, we see that the smoothness penalty when $\text{min}^\gamma$ is used is given by
\begin{align}
d^{\:\gamma} ( \mathbf{a}; \gamma) &:= \text{min}^\gamma(\mathbf{a}; \gamma) - \min(\mathbf{a}), \nonumber \\
 &= - \gamma \; \log \sum_{i=1}^N e^{-(a_i-a_0)/\gamma} \nonumber \\
 &\in [-\gamma \log(N), 0).
\label{eqApp:penaltyMinGamma}
\end{align} 
Therefore, we have shown that $d^{\:\gamma} (\mathbf{a})$ is always negative and achieves its most negative value of $-\gamma \log(N)$ if and only if the input vector $\mathbf{a}$ represents an $N$-way tie, \ie, $\mathbf{a} = 
\begin{bmatrix}a_0& \ldots& a_0\end{bmatrix}^\top$.

\section{Baselines} \label{sec:baselines}
We compare our approach to other weakly supervised \cite{DwibediATSZ19,SermanetLCHJSLB18,ChangHS0N19} and self-supervised \cite{BenaimELMFRID20,MisraZH16} methods that entail temporal understanding in their definition. 

\noindent{\bf SpeedNet} \cite{BenaimELMFRID20} learns a video representation by learning to distinguish between videos played at different speeds.

\noindent{\bf Time Contrastive Networks (TCN)} \cite{SermanetLCHJSLB18} makes fine-grained temporal distinctions by optimizing a contrastive loss that encourages embeddings of an anchor image and an image taken simultaneously from a different camera viewpoint to be similar, while the embedding of an image taken from the same sequence of the anchor but at a different time instant to be distant from that of the anchor.

\noindent{\bf Shuffle and Learn (SaL)} \cite{MisraZH16} learns to predict whether a triplet of frames are in the correct order or shuffled.

\noindent{\bf Discriminative Differentiable Time Warping (D$^3$TW)} \cite{ChangHS0N19} was originally introduced to learn alignments between video frames and action labels. Its definition includes a discriminative component requiring explicit definitions of positive and negative pairs. A pair is considered positive if all action labels are used in the alignment table, whereas negative pairs are constructed by dropping some of the action labels. We adapt this loss to our video pair alignment setting and define \textbf{D$^3$TW*} as follows: a positive pair is constructed by sampling frames from the entire duration of both \textit{video-1} and \textit{video-2} in any pair, whereas we randomly drop portions of \textit{video-2} to constitute a negative pair, thereby mimicking the dropping of action labels as proposed in the original paper.

\noindent{\bf Temporal Cycle Consistency (TCC)} \cite{DwibediATSZ19} learns fine-grained temporal correspondences between individual video frames by imposing a soft version of cycle consistency on the individual matches.

\section{FineGym re-organization} \label{sec:FGR}

\begin{figure}[h]
\begin{center}
\resizebox{0.99\columnwidth}{!}{
	\includegraphics{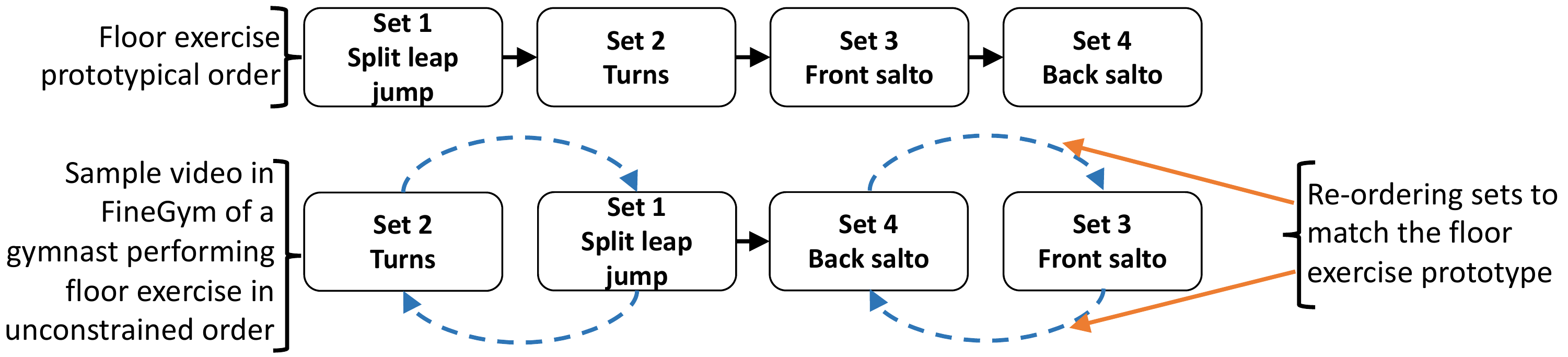}
}
\end{center}
\caption{Illustration of our re-organization of FineGym.}
\label{fig:FineGym-Org}
\vspace{-10pt}
\end{figure}

As mentioned in the main mansucript, each  video  in  FineGym  is annotated according to a three-level hierarchy denoting the event being  performed  in  the  video,  the  different sets involved in performing the event, and the framewise elements (i.e., action phases) involved in each set.  To perform any event-level action, a gymnast may perform the different sets in  any  order.   To  train  our embedding network using  our  alignment-based  method,  we  re-organize  the  FineGym  dataset  such that all sets belonging to the same event appear in the same order in any given video. For example, given floor exercise
events, gymnasts can perform four different sets of exercises
in any order, we re-organize the clips in each video
according to a selected prototype order, as shown Figure \ref{fig:FineGym-Org}. These organized event-level videos are used during training and testing.

\section{Fine-grained action recognition} \label{sec:FGR1}

In Sec.\ 4.4 of the main manuscript, we compared our fine-grained action recognition performance to our baselines using FineGym 
with two training settings for the backbone framewise encoder.  Here, we provide our complete comparison.  In addition, to 
training from scratch (\textbf{train-all}) and fine-tuning the batch norm layers (\textbf{only-bn}), we include
a third experiment consisting of fine-tuning all layers of a ResNet50 model pre-trained on ImageNet (\textbf{train-all}).

The full results of all experiments are summarized in Table \ref{tab:SOTA-comparisoln-full}.  Consistent with our conclusions in 
Sec.\ 4.4, we outperform all the weakly and self-supervised baseline methods by significant margins under all training settings with the best results obtained under the \emph{only-bn} setting. 
\begin{table}[h]
	\begin{center}
	\resizebox{0.9\columnwidth}{!}{
		\begin{tabular}{c|c|c|c}
			\hline
			Method & Training & FineGym101 & FineGym290 \\
			\hline
			SpeedNet \cite{BenaimELMFRID20} & \multirow{6}{*}{scratch} & 30.40& 29.87\\
			TCN \cite{SermanetLCHJSLB18} & & 36.52 & 37.40 \\
			SaL \cite{MisraZH16} & & 40.25 & 37.98 \\
			D$^3$TW* \cite{ChangHS0N19} & & 32.10 & 32.15\\
			TCC \cite{DwibediATSZ19} &  & 41.78 & 40.57 \\
			Ours &  &\textbf{45.79} & \textbf{43.49} \\
			\hline
			SpeedNet \cite{BenaimELMFRID20} &  \multirow{6}{*}{train all} & 28.27 & 30.95\\
			TCN \cite{SermanetLCHJSLB18} & & 30.97 & 35.68 \\
			SaL \cite{MisraZH16} && 36.76 & 39.76 \\
			D$^3$TW* \cite{ChangHS0N19} & & 35.75 & 33.63 \\
			TCC \cite{DwibediATSZ19} &  & 44.29 & 40.31 \\
			Ours &  &\textbf{47.78} & \textbf{44.77} \\
			\hline
			SpeedNet \cite{BenaimELMFRID20} &  \multirow{6}{*}{only bn} & 34.38 & 35.92 \\
			TCN \cite{SermanetLCHJSLB18} & & 41.75 & 39.93 \\
			SaL \cite{MisraZH16} && 42.68 & 41.58 \\
			D$^3$TW* \cite{ChangHS0N19} &  & 38.21 & 34.04 \\
			TCC \cite{DwibediATSZ19} &  & 45.62& 43.40 \\
			Ours & &\textbf{49.51} & \textbf{46.54} \\
			\hline
			
		\end{tabular}
	}
	\end{center}
	\caption{Fine-grained action recognition accuracy on both organizations of FineGym.}
	\label{tab:SOTA-comparisoln-full}
	\vspace{-10pt}
\end{table}

For completeness, we also provide results of training the SVM classifier for framewise fine-grained action recognition on the original FineGym99 and 288 short clips. The results of this experiment, summarized in Table \ref{tab:SOTA-Gymm99-288}, once again demonstrate the superiority of the proposed approach even under this more challenging setting. Notably, comparison between our method and those reported in \cite{shao2020finegym} is not direct for two main reasons. First, we are targeting framewise accuracy, whereas \cite{shao2020finegym} focuses on clip-level accuracy. Second, we are the first to report weakly supervised results on FineGym.

\begin{table}[t]
	\begin{center}
	\resizebox{0.9\columnwidth}{!}{
		\begin{tabular}{c|c|c|c}
			\hline
			Method & Training & FineGym99 & FineGym288 \\
			\hline
			SpeedNet \cite{BenaimELMFRID20} &  \multirow{6}{*}{only bn} & 16.86 & 15.57 \\
			TCN \cite{SermanetLCHJSLB18} &  & 20.02 & 17.11 \\
			SaL \cite{MisraZH16} && 21.45 & 19.58 \\
			D$^3$TW* \cite{ChangHS0N19} &  & 15.28 & 14.07 \\
			TCC \cite{DwibediATSZ19} &  & 25.18 & 20.82 \\
			Ours & &\textbf{27.81} & \textbf{24.16} \\
			\hline
			
		\end{tabular}
	}
	\end{center}
	\caption{Fine-grained action recognition accuracy on the original clips of FineGym99 and FineGym288.}
	\label{tab:SOTA-Gymm99-288}
\end{table}

\section{Details of fine-grained action recognition} \label{sec:FGR2}
To further investigate the utility of the learned embeddings, we also consider classification results of each event separately. In particular, using the embeddings learned on the entire FineGym101 dataset, we train a separate SVM classifier for each event.
The results summarized in Table \ref{tab:detailed-SOTA-comparisons}, further confirm the superiority of our approach. These results also show that classifying the sub-actions in the floor exercise (FX) event is the most challenging for all methods. Careful examination of videos in this class revealed wide variations in the way gymnasts perform each sub-action in the floor exercise event, which makes learning a proper alignment especially challenging.

\begin{table}[h]
	\begin{center}
		\resizebox{0.9\columnwidth}{!}{
		\begin{tabular}{c|c|c|c|c}
			\hline
			Method & VT, \textit{8cls} & FX, \textit{35cls} & BB, \textit{33cls} &  UB, \textit{25cls}\\
			\hline
			SpeedNet \cite{BenaimELMFRID20}  & 61.63 & 12.10 & 24.25 & 24.42\\
			TCN \cite{SermanetLCHJSLB18} & 68.82 & 13.49 & 29.57 & 27.14 \\
			SaL \cite{MisraZH16} & 70.53 &15.16 & 37.73 & 33.06 \\
			D$^3$TW* \cite{ChangHS0N19} &65.79 & 13.57& 31.96& 26.61\\
			TCC \cite{DwibediATSZ19} & 69.21 & 20.01 & 40.95 & 37.08 \\
			Ours & \textbf{70.69} &\textbf{20.06} & \textbf{43.81} & \textbf{40.12}\\
			\hline
			
		\end{tabular}
	}
	\end{center}
	\caption{Detailed evaluation of fine-grained action recognition performance by looking at elements within each event separately.}
	\label{tab:detailed-SOTA-comparisons}
\end{table}

\section{Video synchronization}\label{sec:PennR}

In the main manuscript, we provide results of video synchronization under the challenging setting of training a single network for all classes in PennAction, whereas \cite{DwibediATSZ19} trained a different network for each class. For completeness, we provide additional results here where we also trained a network per class using our loss on PennAction to directly compare with \cite{DwibediATSZ19}. The results summarized in Table \ref{tab:PenR} speak decisively in favor of our approach where we outperform all approaches with a sizeable margin.

\begin{table}[h]
	\begin{center}
		\resizebox{0.5\columnwidth}{!}{
		\begin{tabular}{c|c}
			\hline
			Method & Kendall's Tau\\
			\hline
			TCN \cite{SermanetLCHJSLB18} & 73.28 \\
			SaL \cite{MisraZH16} & 63.36 \\
			TCC \cite{DwibediATSZ19} & 73.53 \\
			Ours & \textbf{78.29} \\
			\hline
			
		\end{tabular}
	}
	\end{center}
	\caption{Video alignment results using Kendall's Tau metric on PennAction. \emph{Consistent} with previous work, these results were obtained by training a separate network for each class separately.}
	\label{tab:PenR}
	\vspace{-10pt}
\end{table}

\begin{figure*}[t]
	\begin{center}
		\resizebox{0.9\textwidth}{!}{
			\begin{tabular}{c}
				\includegraphics{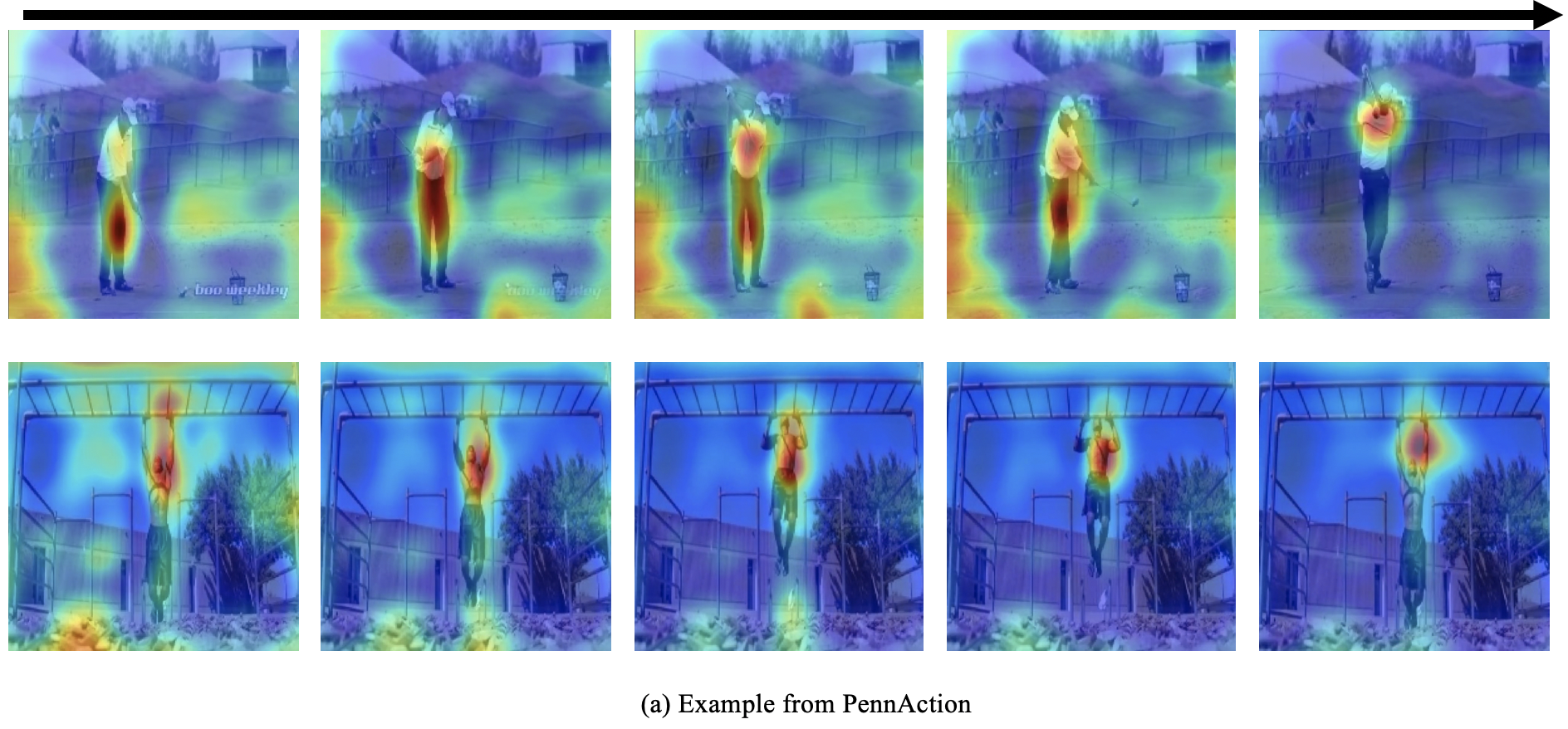} \\
				\includegraphics{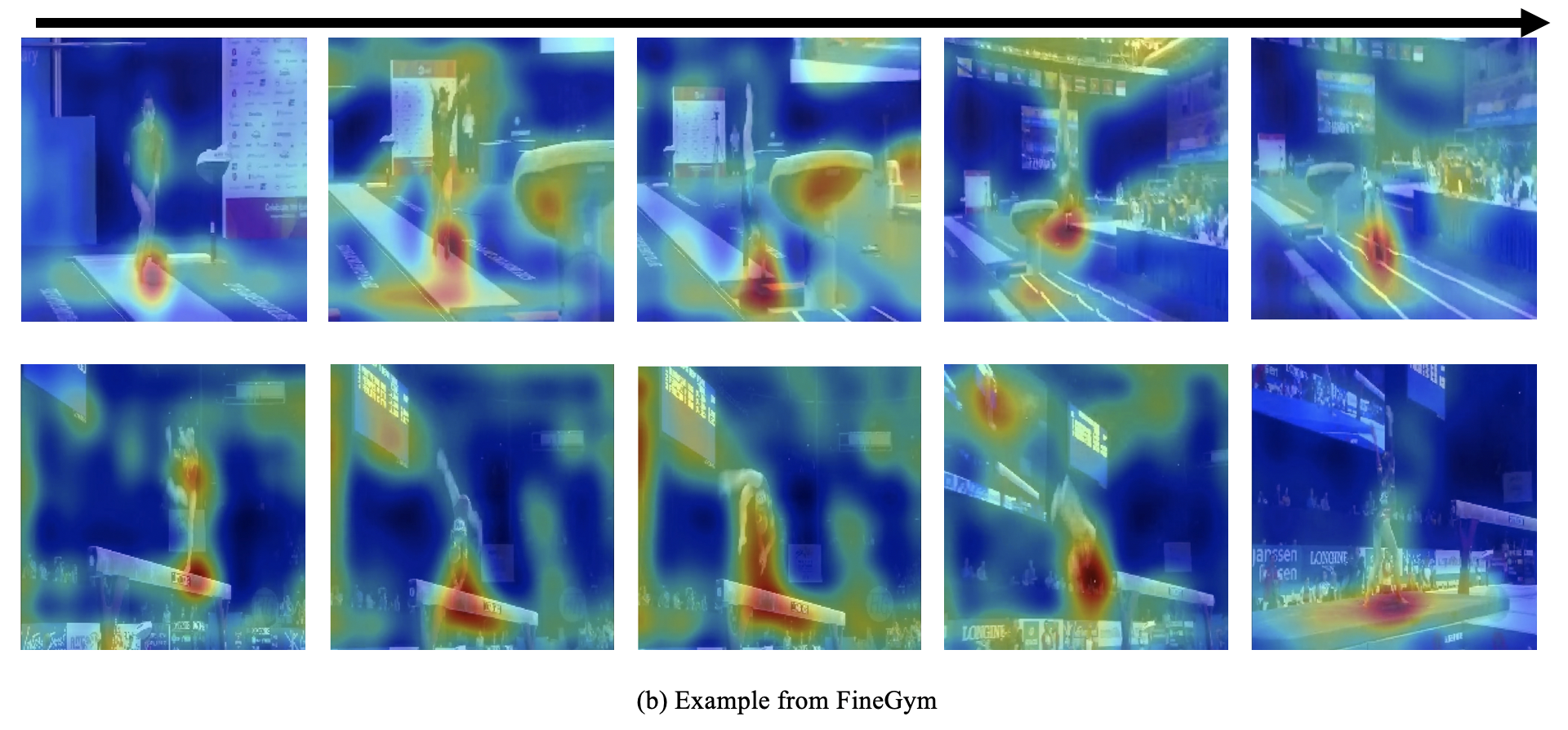} \\
			\end{tabular}
			
		}
	\end{center}
	\caption{Visualization of the features learned with the proposed alignment loss using the CAM method.}
	\label{fig:CAM}
	\vspace{-10pt}
\end{figure*}

\section{Visualizing learned features} \label{sec:visualization}
To investigate what our learned representation captures, we adapt the Class Activation Map (CAM) method \cite{ZhouKLOT16} to visualize the learned features. In particular, we extract feature maps from the last convolutional layer of our embedding network and simply average them along the channel dimension. The resulting activation maps are then normalized between 0 and 1 framewise, upsampled to match the input dimensions, and superimposed on the input video frames. For PennAction, the heatmaps in Figure \ref{fig:CAM} (a) shows that our embeddings are selective to body parts most involved in performing an action. This can explained by the fact that videos in  PennAction  are carefully curated with relatively clean, similar actions with no repetitions. On the other hand, we can see from Figure \ref{fig:CAM} (b), that our embeddings are tuned to human contact with surfaces to learn alignments in FineGym. This is an especially desired behaviour as while there can be significant variations in the way gymnasts perform different phases of an action, they generally share some commonality in the manner that they
make contact with surfaces.
More generally, these visualizations suggest that the proposed loss learns to adapt and identify the most reliable cues to learn the alignments. Additional activation visualizations are provided in the supplemental video.

\section{Downstream applications}
Please see supplemental video \footnote{see video at: https://github.com/hadjisma/VideoAlignment} for various downstream application results.

\end{document}